\documentclass[sigconf,table]{acmart}

\usepackage{subcaption}
\usepackage{graphicx}
\usepackage{booktabs} 
\usepackage{lscape}      
\usepackage{multirow}    
\usepackage{makecell}    
\usepackage{algorithmic} 
\usepackage{algorithm}   
\usepackage{textcomp}    
\usepackage{subcaption}  
\usepackage{tikz}        
\usepackage{soul}        
\usepackage{graphicx}

\AtBeginDocument{%
  }

\setcopyright{acmlicensed}

\renewcommand\footnotetextcopyrightpermission[1]{}
\begin{document}

\title{DOC-GS: Dual-Domain Observation and Calibration for Reliable Sparse-View Gaussian Splatting}

\author{Hantang Li}
\affiliation{
  \institution{Harbin Institute of Technology}
  \city{Shenzhen}
  \country{China}
}
\affiliation{
  \institution{Pengcheng Laboratory}
  \city{Shenzhen}
  \country{China}
}
\email{25B951062@stu.hit.edu.cn}

\author{Qiang Zhu}

\affiliation{
  \institution{Pengcheng Laboratory}
  \city{Shenzhen}
  \country{China}
}
\email{zhuqiang@pcl.ac.cn}

\author{Xiandong Meng}
\authornote{Corresponding authors.}
\affiliation{
  \institution{Pengcheng Laboratory}
  \city{Shenzhen}
  \country{China}
}
\email{mengxd@pcl.ac.cn}

\author{Debin Zhao}
\affiliation{
  \institution{Harbin Institute of Technology}
  \city{Harbin}
  \country{China}
}
\email{dbzhao@hit.edu.cn}

\author{Xiaopeng Fan}
\authornotemark[1]
\affiliation{
  \institution{Harbin Institute of Technology}
  \city{Harbin}
  \country{China}
}
\email{fxp@hit.edu.cn}

\renewcommand{\shortauthors}{Trovato et al.}

\begin{abstract}

Sparse-view reconstruction with 3D Gaussian Splatting (3DGS) is fundamentally ill-posed due to insufficient geometric supervision, often leading to severe overfitting and the emergence of structural distortions and translucent haze-like artifacts. While existing approaches attempt to alleviate this issue via dropout-based regularization, they are largely heuristic and lack a unified understanding of artifact formation. 
In this paper, we revisit sparse-view 3DGS reconstruction from a new perspective and identify the core challenge as the unobservability of Gaussian primitive reliability. 

Unreliable Gaussians are insufficiently constrained during optimization and accumulate as haze-like degradations in rendered images. 
Motivated by this observation, we propose a unified Dual-domain Observation and Calibration (DOC-GS) framework that models and corrects Gaussian reliability through the synergy of optimization-domain inductive bias and observation-domain evidence. 
Specifically, in the optimization domain, we characterize Gaussian reliability by the degree to which each primitive is constrained during training, and instantiate this signal via a Continuous Depth-Guided Dropout (CDGD) strategy, where the dropout probability serves as an explicit proxy for primitive reliability. This imposes a smooth depth-aware inductive bias to suppress weakly constrained Gaussians and improve optimization stability. In the observation domain, we establish a connection between floater artifacts and atmospheric scattering, and leverage the Dark Channel Prior (DCP) as a structural consistency cue to identify and accumulate anomalous regions. 
Based on cross-view aggregated evidence, we further design a reliability-driven geometric pruning strategy to remove low-confidence Gaussians. 
Extensive experiments on multiple benchmarks demonstrate that DOC-GS consistently outperforms existing methods for sparse-view reconstruction, suppressing haze-like artifacts and improving geometric fidelity in three representative datasets.
\end{abstract}

\ccsdesc[500]{Computing methodologies~Reconstruction}
\keywords{Sparse-view Reconstruction; 3D Gaussian Splatting; Dropout; Dark Channel Prior}
\begin{teaserfigure}
\centering
\includegraphics[width=0.99\textwidth]{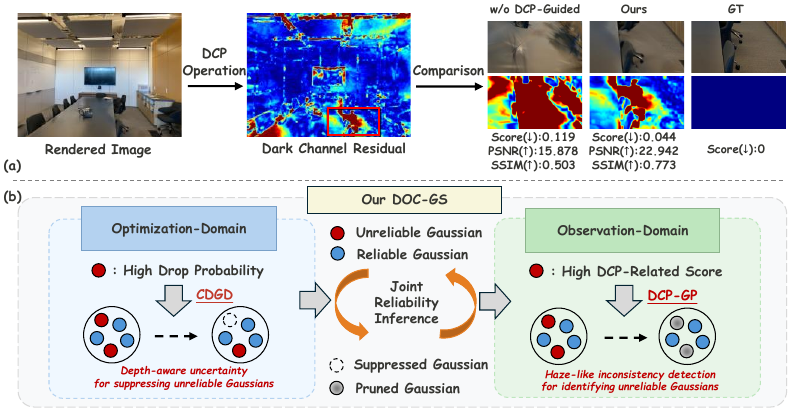}
\caption{Motivation of DOC-GS for sparse-view Gaussian splatting.
(a) To explore the degradation characteristics of rendered images, we generate the dark channel residual between the rendered image and ground-truth image. By comparison on the quantitative and qualitative results, the regions affected by floaters in the rendered images exhibit abnormal dark channel responses, which intuitively highlights artifact-prone regions. 
(b) Based on this insight, we reformulate sparse-view 3D Gaussian splatting as a dual-domain reliability inference problem. DOC-GS first introduces Continuous Depth-Guided Dropout (CDGD) to suppress unreliable Gaussians in the optimization domain, and then designs Dark Channel Prior-guided Geometric Pruning (DCP-GP) to detect and remove haze-like artifacts in the observation domain, ensuring high fidelity sparse-view reconstruction.
}
  \label{fig:teaser}
\end{teaserfigure}


\maketitle

\section{Introduction}

Novel view synthesis (NVS) aims to reconstruct photo-realistic images from unseen viewpoints and has been a long-standing problem in computer vision and graphics~\cite{3dgssurvey,Chatedit-3d,GaussianShader,gsslam,3drsurvey}. Recently, 3D Gaussian Splatting (3DGS)~\cite{3dgs} has demonstrated remarkable performance by combining explicit scene representation with efficient differentiable rendering~\cite{3dgszip,HAC,2dgs,compgs,scaffoldgs,Mip-Splatting}. 
For example, Scaffold-GS~\cite{scaffoldgs} introduces a structured scaffold representation to regularize Gaussian distribution, improving rendering fidelity and stability in complex scenes. However, their performance degrades significantly under sparse-view settings, where insufficient geometric constraints lead to severe overfitting, resulting in structural distortions and pervasive translucent floater artifacts~\cite{InstantSplat, nerfand3dgs,COLMAP-Free,Binocular-Guided3DGaussianSplatting,SplatField,Cor-GS}.
Moreover, some works attempt to alleviate this issue by introducing dropout~\cite{Dropout}-based regularization. For example, DropAnSH-GS~\cite{DropAnSHGS} enhances regularization by discarding clusters of neighboring Gaussians and applying Dropout to spherical harmonic coefficients. These methods randomly suppress a subset of Gaussian primitives or design heuristic masking strategies based on spatial cues~\cite{DropoutGS,DropGaussian}. While effective to some extent, they fundamentally operate as stochastic perturbations and lack a principled understanding of which Gaussian primitives are reliable. Additionally, due to the inherent redundancy and neighborhood compensation effect in 3DGS, suppressing individual Gaussians can be easily compensated by nearby primitives, making such strategies insufficient to eliminate artifacts.

In this work, we revisit sparse-view 3DGS from a new perspective. We identify that the core challenge lies in the unobservability of Gaussian primitive reliability. Under sparse supervision, multiple Gaussian configurations can produce nearly identical renderings on training views, making it inherently ambiguous to determine which primitives correspond to true scene structures. As a result, unreliable Gaussians, driven by local optimization errors rather than geometric consistency, persist during training and accumulate into structured artifacts. Interestingly, these artifacts are not random noise. Instead, they exhibit consistent spatial patterns: low-contrast, semi-transparent regions that closely resemble haze effects in atmospheric scattering. As illustrated in Fig.~\ref{fig:teaser} (a), this observation suggests that the degradation is not only an optimization issue but also a phenomenon that can be characterized in the image domain.
Based on these insights, we reformulate sparse-view 3DGS as a dual-domain reliability inference problem. The reliability of Gaussian primitives is jointly inferred from two complementary domains: (1) the optimization domain, which reflects how Gaussians are constrained during training, and (2) the observation domain, which captures how they manifest in rendered images.

To this end, we propose \textbf{D}ual-domain \textbf{O}bservation and \textbf{C}alibration (\textbf{DOC-GS}) framework for reliable \textbf{G}aussian \textbf{S}platting. The pipeline of our DOC-GS is illustrated in Fig.~\ref{fig:teaser} (b). DOC-GS consists of two synergistic components. First, in the optimization domain, we introduce a Continuous Depth-Guided Dropout (CDGD) strategy that imposes a smooth depth-aware inductive bias, suppressing weakly constrained Gaussians and stabilizing the optimization process. Second, in the observation domain, we establish a connection between floater artifacts and atmospheric scattering, and leverage the Dark Channel Prior (DCP) ~\cite{dcp} as a structural consistency cue to detect anomalous regions in rendered images. By aggregating such evidence across views, we design a reliability-driven geometric pruning strategy to remove persistently unreliable Gaussian primitives. 
The main contributions are summarized as follows:

\begin{itemize}
\item We reformulate sparse-view 3DGS reconstruction as a dual-domain reliability inference problem, providing a unified explanation of artifact formation from both optimization and observation perspectives.
\item We propose a continuous depth-guided dropout mechanism that introduces a smooth inductive bias for suppressing unreliable Gaussian primitives.
\item We establish a novel connection between floater artifacts and atmospheric scattering model, and introduce the DCP as an observation-domain cue for structural anomaly detection. A reliability-driven geometric pruning strategy is designed based on cross-view evidence accumulation to effectively remove low-confidence Gaussians.
\item Extensive experiments demonstrate that our DOC-GS significantly improves rendering quality and reduces haze-like artifacts under all sparse-view settings for three widely-used datasets.
\end{itemize}

\section{Related Work}

\subsection{Novel View Synthesis with 3D Gaussian Splatting}

Novel View Synthesis (NVS) aims to render photorealistic images from unseen viewpoints given a set of input observations. Neural Radiance Fields (NeRF)~\cite{Atzmon2019,Mip-NeRF,AutoInt,Michalkiewicz2019,NeRF,Muller2022,Niemeyer2019,DeepSDF,Peng2020,Sun2022,Zhang2022,FangOneIsAll2023,Garbin2021,Sharp-NeRF,MartinBrualla2021,KiloNeRF,UAVNeRF,PointNeRF} represent a seminal paradigm by modeling scenes as continuous volumetric functions, achieving high fidelity at the cost of high computational overhead.
Recently, 3D Gaussian Splatting (3DGS)~\cite{GaussianShader,3dgs,scaffoldgs,SplatField} has emerged as an efficient alternative based on explicit representations. By modeling scenes as anisotropic Gaussian primitives and leveraging differentiable rasterization with depth-aware $\alpha$-blending, 3DGS achieves real-time rendering while maintaining competitive visual quality~\cite{Das2024,LeeJooChan2024,Mip-Splatting}. However, its performance degrades significantly under sparse-view settings due to insufficient geometric constraints.

\subsection{Novel View Synthesis with Sparse View}

Sparse-view NVS aims to reconstruct novel views from limited observations, posing a highly ill-posed problem. Prior works address this challenge through additional supervision or structural constraints. NeRF-based approaches introduce depth supervision, semantic consistency, or multi-view regularization~\cite{Mip-NeRF,COLMAP-Free,Jain2021,SplatField,RegNeRF,SparseNeRF,FreeNeRF}.
In the context of 3DGS, several methods aim to improve geometric consistency. FSGS~\cite{FSGS} introduces proximity-based Gaussian expansion, while CoR-GS~\cite{Cor-GS} enforces collaborative regularization across models. DNGaussian~\cite{DNGaussian} and NexusGS~\cite{NexusGS} incorporate depth normalization and epipolar constraints. More recently, D$^2$GS~\cite{D2GS} identifies spatial imbalance in sparse-view reconstruction and proposes depth-aware regularization to address overfitting in near regions and under-constrained far regions.
Despite these advances, existing approaches primarily focus on improving optimization stability and do not explicitly model the structural reliability of Gaussian primitives.

\subsection{Optimization-domain Regularization via  Dropout}

Inspired by regularization strategies in deep learning, recent works introduce dropout mechanisms into sparse-view 3DGS to mitigate overfitting. DropGaussian~\cite{DropGaussian} and DropoutGS~\cite{DropoutGS} randomly suppress Gaussian primitives during training to reduce over-reliance on individual components.
However, such stochastic strategies are inherently prior-free and lack structural awareness. Due to the redundancy of Gaussian representations, removing individual primitives can be compensated by neighboring Gaussians, leading to the neighborhood compensation effect~\cite{DropGaussian,CoAdaptation,DropAnSHGS}. This results in limited changes in rendered outputs and weak supervision signals.
To address this issue, more structured dropout strategies have been proposed. DropAnSH-GS~\cite{DropAnSHGS} introduces anchor-based group dropout, while D$^2$GS~\cite{D2GS} incorporates depth and density cues for adaptive suppression. Nevertheless, these methods rely on discrete masking operations and do not explicitly characterize the reliability of Gaussian primitives.

\subsection{Observation-domain Priors}

Image-space priors have been extensively studied in low-level vision. The Atmospheric Scattering Model (ASM)~\cite{acm1,acm2} describes image formation under participating media, while the Dark Channel Prior (DCP)~\cite{dcp} exploits statistical properties of haze-free images for single-image dehazing. Despite their success in image restoration, their integration into 3D reconstruction remains largely unexplored. 
In sparse-view 3D Gaussian Splatting, rendering artifacts exhibit structured degradations that are consistent with haze-like statistical patterns in the image space, indicating that image-space priors can provide informative cues for identifying structural inconsistencies. However, existing methods largely overlook such observation-domain information during the reconstruction process.

\subsection{Discussion}

Existing methods primarily operate in either the optimization domain or the observation domain. Optimization-based approaches introduce stochastic or heuristic regularization but lack explicit modeling of structural reliability, while image-based priors capture artifact patterns but are not integrated into the reconstruction process.
In contrast, we formulate sparse-view 3DGS as a dual-domain reliability inference problem, where Gaussian reliability is jointly estimated from optimization dynamics and image observations. This perspective unifies existing approaches and provides a principled framework for suppressing unreliable Gaussian primitives.

\begin{figure*}[t]
    \centering
    \includegraphics[width=\textwidth]{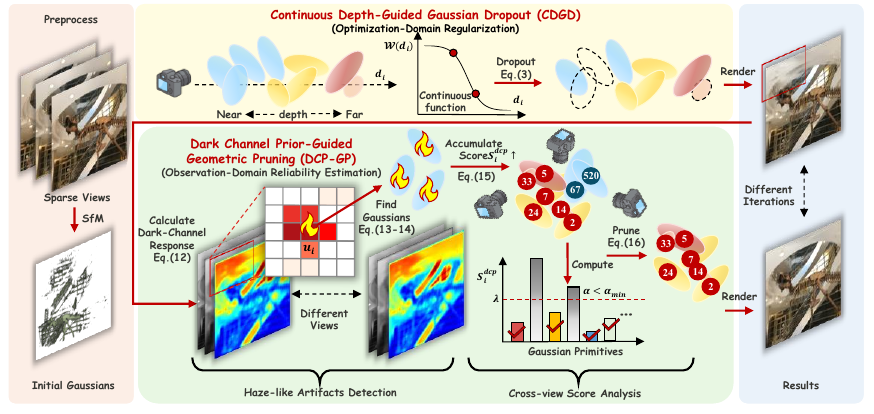}
    \caption{Overview of DOC-GS. 
    Initial Gaussian primitives from sparse-view SfM are jointly optimized through two complementary domains. In the optimization domain, Continuous Depth-Guided Dropout (CDGD) introduces depth-aware uncertainty to suppress potentially unreliable Gaussians, improving optimization stability. 
    In the observation domain, we leverage the Dark Channel Prior (DCP) to detect haze-like structural inconsistencies in rendered images. The resulting evidence is accumulated across views to estimate Gaussian reliability, which guides a reliability-driven geometric pruning strategy to remove persistently unreliable primitives. 
    The entire process is performed iteratively, enabling progressive refinement of Gaussian representations and effective suppression of haze-like artifacts under sparse-view settings.
    }
    \label{fig:framework}
\end{figure*}

\section{Methodology}
\label{sec:method}

\subsection{Dual-Domain Perspective}

Sparse-view 3DGS is inherently under-constrained, making it difficult to determine which Gaussian primitives are reliable. We address this issue from a dual-domain perspective: the optimization domain provides depth-dependent regularization signals during training, while the observation domain reveals anomaly patterns in rendered images. As shown in Fig.~\ref{fig:framework}, our method integrates these two complementary signals through dropout-based regularization and DCP-guided pruning to progressively refine the reliable Gaussian representation.

\subsection{Continuous Depth-Guided Gaussian Dropout (CDGD)}
\label{subsec:continuous_dropout}

Prior work ~\cite{D2GS} shows that depth serves as an effective proxy for spatial imbalance, where primitives at different depths exhibit different tendencies toward overfitting or insufficient constraint. According to this, in the optimization domain, we regulate Gaussian primitives according to their depth-dependent characteristics during training and the dropout probability is defined via a piecewise step function:
\begin{equation}
    P_i = 
    \begin{cases} 
    D_i, & d_i \le D_{near}, \\ 
    \lambda_{middle} D_i, & D_{near} < d_i \le D_{middle}, \\ 
    \lambda_{far} D_i, & d_i > D_{middle} ,
    \end{cases}
\end{equation}
where $d_i$ represents the camera-space depth of the $i$-th Gaussian primitive, $D_i$ denotes the normalized depth-based importance score obtained via min-max normalization, and $P_i$ is the corresponding dropout probability. $D_{near}$ and $D_{middle}$ define the depth thresholds for partitioning the scene into different regions, while $\lambda_{middle}$ and $\lambda_{far}$ are scaling factors that modulate the dropout probability in the middle and far regions, respectively.

However, this discrete binning introduces non-differentiable transitions in 3D space, which may lead to unstable optimization and disrupt local structural coherence. To obtain a smoother and more stable regularization signal, we replace the hard binning strategy with a continuous depth-guided weighting function:
\begin{equation}
    \mathcal{W}(d_i) = \lambda_{base} + \frac{1 - \lambda_{base}}{1 + \exp\left(\kappa (d_i - \tau)\right)},
\end{equation}
where $\mathcal{W}(d_i)$ denotes the depth-dependent attenuation weight, $\lambda_{base} \in (0,1]$ defines the lower bound of the weighting function, $\tau$ represents the transition center (set as the median depth of the scene), and $\kappa$ controls the steepness of the transition. Therefore, the final dropout probability is given by:
\begin{equation}
    P_i = D_i \cdot \mathcal{W}(d_i),
\end{equation}
where $P_i$ denotes the continuous dropout probability of the $i$-th Gaussian primitive.
This formulation provides a continuous and stable optimization-domain signal, enabling smooth spatial regularization without introducing abrupt perturbations. It helps suppress unreliable Gaussian primitives during training while preserving structural consistency.

\subsection{From Floater Artifacts to an Analogous Atmospheric Scattering Interpretation}

In the observation domain, unreliable Gaussians often manifest as semi-transparent and spatially disordered floaters in rendered images. To better characterize this phenomenon, we analyze how such floaters affect the image formation process in 3DGS. Recall the standard volume rendering formulation of 3DGS:
\begin{equation}
    C = \sum_{i=1}^{N} c_i \alpha_i \prod_{j=1}^{i-1} (1 - \alpha_j),
\end{equation}
where $C$ represents the final rendered color of a pixel, $N$ denotes the number of Gaussian primitives intersecting the corresponding camera ray, $c_i$ represents the color of the $i$-th Gaussian primitive, and $\alpha_i$ denotes its opacity. The product term $\prod_{j=1}^{i-1} (1 - \alpha_j)$ represents the accumulated transmittance of all preceding Gaussians along the ray.
Consider a camera ray that first passes through a region dominated by floating artifacts before reaching the actual scene surface. Let the first $K$ Gaussians along the ray correspond to floaters, while the remaining primitives account for the true scene content. The rendering equation can then be decomposed as:
\begin{equation}
    C = C_F + T_F C_{\text{surf}},
\end{equation}
where $C_F$ represents the aggregated color contribution of floater-related Gaussians, $T_F$ denotes the accumulated transmittance through the floater region, and $C_{\text{surf}}$ represents the radiance contribution of the underlying true surface after passing through the floaters.
More specifically, $C_F$ and $T_F$ are defined as:
\begin{equation}
    C_F = \sum_{i=1}^{K} c_i \alpha_i \prod_{j=1}^{i-1} (1 - \alpha_j), 
    \quad
    T_F = \prod_{j=1}^{K} (1 - \alpha_j).
\end{equation}

To further understand the behavior of $C_F$, we consider the common sparse-view case in which floaters appear as a dense set of low-opacity Gaussian elements distributed in free space. Under mild statistical assumptions, namely: (i) the color statistics of these floaters are locally stationary, and (ii) their colors and opacities are weakly correlated, the aggregated contribution of the floater layer can be approximated as:
\begin{equation}
    C_F \approx A (1 - T_F),
\end{equation}
where $A$ represents the expected radiance of the floater field, which can be interpreted as a low-frequency ambient color induced by the accumulated floater layer, and $1-T_F$ reflects the effective attenuation caused by the floaters.
Substituting the above approximation into the rendering decomposition yields:
\begin{equation}
    C \approx A (1 - T_F) + C_{\text{surf}} T_F.
    \label{eq:floater_approx}
\end{equation}
This formulation bears an \emph{analogous} form to the atmospheric scattering model:
\begin{equation}
    I(\mathbf{u}) = A \bigl(1 - t(\mathbf{u})\bigr) + J(\mathbf{u}) t(\mathbf{u}),
\end{equation}
where $I(\mathbf{u})$ represents the observed hazy image at pixel coordinate $\mathbf{u}$, $A$ denotes the atmospheric light, $t(\mathbf{u})$ is the transmission map, and $J(\mathbf{u})$ represents the haze-free scene radiance.

We emphasize that the above derivation is introduced only as an interpretive analogy, rather than a claim that floater artifacts obey the same physical image formation process as real atmospheric scattering. Nevertheless, the comparison suggests that floater-induced degradations share two important characteristics with haze: they introduce an additional low-frequency veil-like component and attenuate the contribution of the underlying scene structure. This observation motivates us to use dark-channel-based image statistics as an observation-domain cue for detecting such artifacts.

\subsection{Observation-Domain Reliability Estimation via DCP-Guided Geometric Pruning (DCP-GP)}

Based on the above observation-domain interpretation, we introduce a Dark Channel Prior (DCP)-inspired strategy to detect and accumulate anomaly evidence associated with floater artifacts. Although floater-induced degradations are not physically equivalent to atmospheric haze, their appearance often produces abnormally elevated dark-channel responses in rendered images. We therefore use dark-channel statistics as an observation-domain cue for identifying structurally inconsistent renderings.

Let $\mathcal{D}(\cdot)$ denote a dark-channel-inspired operator that computes the minimum response across color channels, optionally followed by local aggregation. Applying it to the floater-aware rendering formulation in Eq.~\eqref{eq:floater_approx} yields:
\begin{equation}
    \mathcal{D}(C)(\mathbf{u}) = \mathcal{D}\Bigl( A \bigl(1 - T_F(\mathbf{u})\bigr) + C_{\text{surf}}(\mathbf{u}) T_F(\mathbf{u}) \Bigr),
\end{equation}
where $\mathbf{u} \in \mathbb{R}^2$ represents the pixel coordinate on the image plane, and $\mathcal{D}(C)(\mathbf{u})$ denotes the dark-channel response of the rendered image.
Due to the nonlinearity of $\mathcal{D}(\cdot)$, an exact analytical decomposition is intractable. Instead, we introduce three mild assumptions: (i) the dark channel of the clean surface component is inherently small ($\mathcal{D}(C_{\text{surf}}) \approx 0$), (ii) the floater-induced radiance $A$ varies smoothly within local neighborhoods, and (iii) floater-dominated regions correspond to relatively low transmittance. Under these conditions, the dark-channel response can be approximately attributed to the veil-like floater component:
\begin{equation}
    \mathcal{D}(C)(\mathbf{u}) \approx \mathcal{D}\bigl(A (1 - T_F(\mathbf{u}))\bigr) \approx \mathcal{D}(A)\bigl(1 - T_F(\mathbf{u})\bigr),
\end{equation}
where $\mathcal{D}(A)$ denotes the local dark-channel response of the smooth floater radiance term. This approximation suggests that elevated dark-channel responses can serve as a coarse indicator of accumulated floater opacity along each camera ray.
We emphasize that this formulation does not strictly follow the classical dark channel prior, which relies on a patch-wise minimum operator. Instead, we adopt a simplified variant combined with local averaging, which is more stable for rendered images and less sensitive to noise under sparse-view supervision, while still preserving the low-intensity sensitivity characteristic of dark-channel statistics.
However, directly pruning Gaussians based on a single observation is unreliable, as naturally dark regions or shadows may also produce strong responses. To improve robustness, we adopt a multi-view accumulation strategy and activate DCP-based monitoring after initial densification stage.

Specifically, for each rendered image, we compute the pixel-wise minimum across color channels:
\begin{equation}
    D(\mathbf{u}) = \min_{c \in \{r,g,b\}} C^c(\mathbf{u}).
\end{equation}
We further compute a locally averaged response $\tilde{D}(\mathbf{u})$ to incorporate spatial context. A pixel is marked as anomalous if it violates both local and pixel-level thresholds:
\begin{equation}
    \mathbf{1}_{\text{bad}}(\mathbf{u}) = (\tilde{D}(\mathbf{u}) > \tau_1) \land (D(\mathbf{u}) > \tau_2),
\end{equation}
where $\mathbf{1}_{\text{bad}}(\mathbf{u})$ is a binary indicator, and $\tau_1, \tau_2$ are predefined thresholds.
The overall anomaly level of the current view is then summarized by a global violation ratio:
\begin{equation}
    r^{\text{dcp}} = \frac{1}{|\mathcal{U}|} \sum_{\mathbf{u} \in \mathcal{U}} \mathbf{1}_{\text{bad}}(\mathbf{u}),
\end{equation}
where $|\mathcal{U}|$ represents the total number of pixels in the image domain $\mathcal{U}$.
Instead of constructing explicit pixel-to-Gaussian correspondences, which are notoriously unstable under sparse-view supervision, we adopt a view-level accumulation strategy:
\begin{equation}
    S_i^{\text{dcp}} \leftarrow S_i^{\text{dcp}} + r^{\text{dcp}}, \quad \forall G_i \in \mathcal{G}_{\text{vis}},
\end{equation}
where $S_i^{\text{dcp}}$ denotes the accumulated anomaly score of the $i$-th Gaussian primitive $G_i$, and $\mathcal{G}_{\text{vis}}$ denotes the set of Gaussians visible in the current view.
Based on the accumulated evidence, we perform periodic pruning:
\begin{equation}
    \text{Prune}(i) = (S_i^{\text{dcp}} > \lambda) \land (\alpha_i < \alpha_{\min}),
\end{equation}
where $\alpha_i$ is the opacity of $G_i$, $\alpha_{\min}$ is the minimum opacity threshold, and $\lambda$ is the dynamic pruning threshold. Following our optimization schedule, $\lambda$ is defined as:
\begin{equation}
    \lambda = \eta \cdot T_{\text{prune}},
\end{equation}
where $\eta$ is a scaling factor controlling pruning sensitivity, and $T_{\text{prune}}$ denotes the pruning interval.

Through this process, Gaussian primitives that persistently contribute to observation-domain anomalies are progressively removed. This observation-domain pruning complements the optimization-domain regularization, together forming a unified dual-domain calibration framework for sparse-view reconstruction.

\begin{table*}[t]
\centering
\caption{Quantitative comparison on LLFF dataset under different sparse-view settings (3, 6, and 9 views). Colors indicate \colorbox{red!30}{best}, \colorbox{orange!30}{second best}, and \colorbox{yellow!30}{third best} results.}
\label{tab:llff_results}
\small 
\setlength{\tabcolsep}{5.5pt} 
\begin{tabular}{ll ccc | ccc | ccc}
\toprule
&  \multirow{2}{*}{Methods} & \multicolumn{3}{c}{3-view} & \multicolumn{3}{c}{6-view} & \multicolumn{3}{c}{9-view} \\
\cmidrule(lr){3-5} \cmidrule(lr){6-8} \cmidrule(lr){9-11}
&  & PSNR$\uparrow$ & SSIM$\uparrow$ & LPIPS$\downarrow$ & PSNR$\uparrow$ & SSIM$\uparrow$ & LPIPS$\downarrow$ & PSNR$\uparrow$ & SSIM$\uparrow$ & LPIPS$\downarrow$ \\
\midrule
\multirow [c]{5}{*}{\rotatebox{0}{NeRF-based}} 
&  Mip-NeRF~\cite{Mip-NeRF} & 16.11 & 0.401 & 0.460 & 22.91 & 0.756 & 0.213 & 24.88 & 0.826 & 0.170 \\
&  DietNeRF~\cite{Jain2021}  & 14.94 & 0.370 & 0.496 & 21.75 & 0.717 & 0.248 & 24.28 & 0.801 & 0.183 \\
&  RegNeRF~\cite{RegNeRF}  & 19.08 & 0.587 & 0.336 & 23.10 & 0.760 & 0.206 & 24.86 & 0.820 & 0.161 \\
&  FreeNeRF ~\cite{FreeNeRF} & 19.63 & 0.612 & 0.308 & 23.73 & 0.779 & 0.195 & 25.13 & 0.827 & 0.160 \\
&  SparseNeRF ~\cite{SparseNeRF} & 19.86 & 0.624 & 0.328 & -- & -- & -- & -- & -- & -- \\
\midrule
\multirow [c]{7}{*}{\rotatebox{0}{3DGS-based}} 
&  3DGS ~\cite{3dgs} & 19.22 & 0.649 & 0.229 & 23.80 & 0.814 & 0.125 & 25.44 & 0.860 & 0.096 \\
&  DNGaussian ~\cite{DNGaussian} & 19.12 & 0.591 & 0.294 & 22.18 & 0.755 & 0.198 & 23.17 & 0.788 & 0.180 \\
&  FSGS ~\cite{FSGS} & 20.43 & 0.682 & 0.248 & 24.09 & \cellcolor{yellow!30}0.823 & 0.145 & 25.31 & \cellcolor{yellow!30}0.860 & 0.122 \\
&  CoR-GS~\cite{Cor-GS} & 20.45 & 0.712 & \cellcolor{yellow!30}0.196 & 24.49 & \cellcolor{orange!30}0.837 & \cellcolor{orange!30}0.115 & \cellcolor{yellow!30}26.06 & \cellcolor{orange!30}0.874 & \cellcolor{yellow!30}0.089 \\
&  DropGaussian ~\cite{DropGaussian} & \cellcolor{yellow!30}20.76 & \cellcolor{yellow!30}0.713 & 0.200 & 24.74 & \cellcolor{orange!30}0.837 & \cellcolor{yellow!30}0.117 & \cellcolor{orange!30}26.21 & \cellcolor{orange!30}0.874 & \cellcolor{orange!30}0.088 \\
& NexuGS~\cite{NexusGS} & \cellcolor{orange!30}21.07 & \cellcolor{orange!30}0.738 & \cellcolor{orange!30}0.177 & -& - & - & - & \c- & - \\ 
&  \textbf{DOC-GS (Ours)} & \cellcolor{red!30}21.38 & \cellcolor{red!30}0.748 & \cellcolor{red!30}0.176 & \cellcolor{red!30}24.88 & \cellcolor{red!30}0.840 & \cellcolor{red!30}0.112 & \cellcolor{red!30}26.23 & \cellcolor{red!30}0.876 & \cellcolor{red!30}0.087 \\
\bottomrule
\end{tabular}
\end{table*}

\subsection{Loss Functions}
\label{subsec:loss}

We adopt the standard 3DGS photometric loss, which minimizes the discrepancy between rendered images and ground-truth images using a combination of L1 loss and SSIM loss.

\begin{equation}
\mathcal{L} = 
\mathcal{L}_{L_1}(I, I_{\text{gt}}) + \lambda_1 \left(1 - \text{SSIM}(I, I_{\text{gt}})\right),
\end{equation}
where $I$ and $I_{\text{gt}}$ denote the rendered image and the ground-truth image, respectively, $\mathcal{L}_{L_1}$ represents the pixel-wise L1 loss, $\text{SSIM}(\cdot)$ denotes the structural similarity index, and $\lambda_1$ is a balancing weight that controls the trade-off between the two terms.



\section{Experiments}

\subsection{Implementation Details}
\noindent\textbf{Datasets and Metrics.} Following previous methods~\cite{DropGaussian,DropoutGS}, our experiments are conducted on three widely-used standard datasets: two real-world datasets, i.e., LLFF~\cite{Mildenhall2019} and MipNeRF-360~\cite{Barron2022}, and one synthetic dataset, i.e., Blender~\cite{NeRF}. We follow the experimental setup of previous work~\cite{DropGaussian,FreeNeRF}, adopting the same data segmentation and downsampling procedures. Additionally,  we employ three widely-used evaluation metrics for evaluating the rendering quality: Peak Signal-to-Noise Ratio (PSNR), Structural Similarity Index (SSIM), and Learned Perceptual Image Patch Similarity (LPIPS).


\begin{table}[t]
\centering
\caption{Quantitative comparison on MipNeRF-360 dataset under different sparse-view setting (12 and 24 views). 
}
\label{tab:mipnerf360_results}
\small
\setlength{\tabcolsep}{3pt} 
\begin{tabular}{l ccc | ccc}
\toprule
\multirow{2}{*}{Methods} & \multicolumn{3}{c}{12-view} & \multicolumn{3}{c}{24-view} \\
\cmidrule(lr){2-4} \cmidrule(lr){5-7}
& PSNR$\uparrow$ & SSIM$\uparrow$ & LPIPS$\downarrow$ & PSNR$\uparrow$ & SSIM$\uparrow$ & LPIPS$\downarrow$ \\
\midrule
3DGS ~\cite{3dgs} & 18.52 & 0.523 & 0.415 & 22.80 & 0.708 & 0.276 \\
FSGS ~\cite{FSGS} & 18.80 & 0.531 & 0.418 & 23.70 & 0.745 & 0.230 \\
CoR-GS ~\cite{Cor-GS} & \cellcolor{yellow!30}19.52 & 0.558 & 0.418 & 23.39 & 0.727 & 0.271 \\
DropGaussian ~\cite{DropGaussian} & \cellcolor{orange!30}19.74 & \cellcolor{yellow!30}0.577 & \cellcolor{yellow!30}0.364 & \cellcolor{yellow!30}23.75 & \cellcolor{orange!30}0.756 & \cellcolor{yellow!30}0.227 \\
NexusGS ~\cite{NexusGS} & - & - & - & \cellcolor{orange!30}23.86 & \cellcolor{yellow!30}0.753 & \cellcolor{red!30}0.206 \\
\textbf{DOC-GS (Ours)} & \cellcolor{red!30}20.11 & \cellcolor{red!30}0.589 & \cellcolor{red!30}0.354 & \cellcolor{red!30}24.17 & \cellcolor{red!30}0.765 & \cellcolor{orange!30}0.212 \\
\bottomrule
\end{tabular}
\end{table}

\noindent\textbf{State-of-the-art methods.} We compare our method with several state-of-the-art approaches, including NeRF-based methods (Mip-NeRF~\cite{Mip-NeRF}, DietNeRF~\cite{Jain2021}, RegNeRF~\cite{RegNeRF}, FreeNeRF~\cite{FreeNeRF}, SparseNeRF~\cite{SparseNeRF}) and 3DGS-based methods (3DGS~\cite{3dgs}, DNGaussian~\cite{DNGaussian}, FSGS~\cite{FSGS}, CoR-GS~\cite{Cor-GS}, NexusGS~\cite{NexusGS}, etc.). Furthermore, we pay particular attention to comparisons with several Dropout-based methods, including DropGaussian~\cite{DropGaussian} and DropoutGS~\cite{DropoutGS}.

\noindent\textbf{Training Details.}
Our method is built upon the standard 3DGS framework with the original optimizer and learning rate schedule unchanged for fair comparison. All scenes are trained for 10,000 iterations.
For CDGD, the transition center $\tau$ is adaptively updated based on scene-specific depth statistics, while the steepness parameter is fixed to $\kappa = 10$ to ensure a smooth yet effective transition.
For DCP-GP, we apply a warm-up stage of $t_{\text{start}} = 5000$ iterations before activating pruning. Afterward, pruning is performed every $T_{\text{prune}} = 1000$ iterations, aligned with the densification schedule. The pruning threshold is set as $\lambda = 0.5 \cdot T_{\text{prune}}$ to maintain a balanced pruning strength.
The dark channel thresholds are fixed to $\tau_1 = 0.10$ and $\tau_2 = 0.05$, determined from the 95-th percentile statistics in early training. The opacity threshold is set to $\alpha_{\min} = 0.05$ to target low-opacity floater Gaussians. All experiments are implemented in PyTorch and trained on a single NVIDIA A800 GPU.

\begin{table}[t]
\centering
\caption{Quantitative comparison on Blender dataset under sparse-view setting  (8 views). 
}
\label{tab:blender_results}
\small
\setlength{\tabcolsep}{12pt}
\begin{tabular}{l c c c c}
\toprule
 Methods & PSNR$\uparrow$ & SSIM$\uparrow$ & LPIPS$\downarrow$ \\


\midrule

3DGS~\cite{3dgs}                    & 21.56 & 0.847 & 0.130 \\
DNGaussian~\cite{DNGaussian}        & 24.31 & 0.886 & 0.088 \\
NexusGS~\cite{NexusGS}       & 24.37 & 0.893 & \cellcolor{yellow!30}0.087\\
FSGS~\cite{FSGS}                    & \cellcolor{yellow!30}24.64 & \cellcolor{yellow!30}0.895 & 0.095 \\
CoR-GS~\cite{Cor-GS}                 & 24.43 &\cellcolor{orange!30}0.896 & \cellcolor{orange!30}0.084 \\
DropGaussian~\cite{DropGaussian}    & \cellcolor{orange!30}25.42& 0.888 & 0.089 \\
\textbf{DOC-GS (Ours)}                       & \cellcolor{red!30}25.61 &\cellcolor{red!30}0.898 & \cellcolor{red!30}0.082 \\

\bottomrule
\end{tabular}
\end{table}


\begin{figure*}[!t]
    \centering
    \small 
    \begin{minipage}{0.19\textwidth}
        \centering
        \text{3DGS} \\ \vspace{1mm}
        \includegraphics[width=\textwidth]{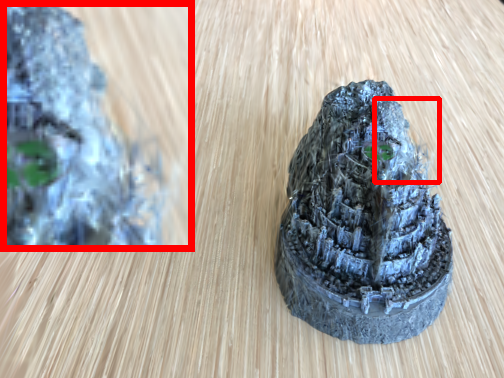} \\ \vspace{1mm}
        \includegraphics[width=\textwidth]{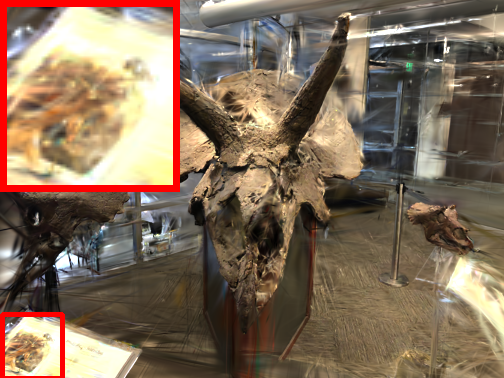} \\ \vspace{1mm}
        \includegraphics[width=\textwidth]{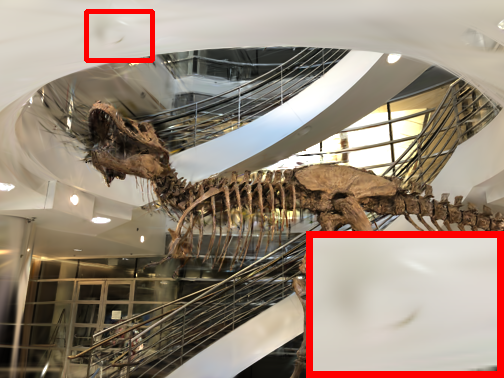}
    \end{minipage}
    \hfill
    \begin{minipage}{0.19\textwidth}
        \centering
        \text{DropoutGS} \\ \vspace{1mm}
        \includegraphics[width=\textwidth]{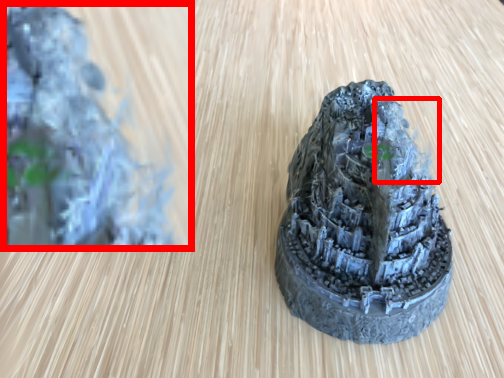} \\ \vspace{1mm}
        \includegraphics[width=\textwidth]{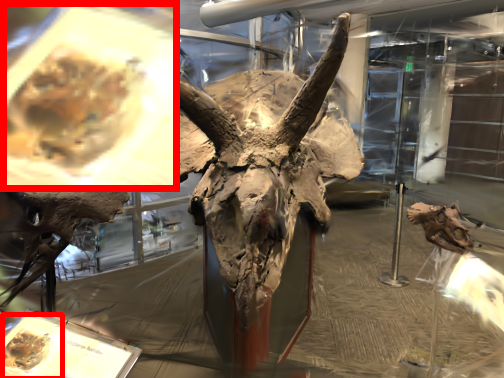} \\ \vspace{1mm}
        \includegraphics[width=\textwidth]{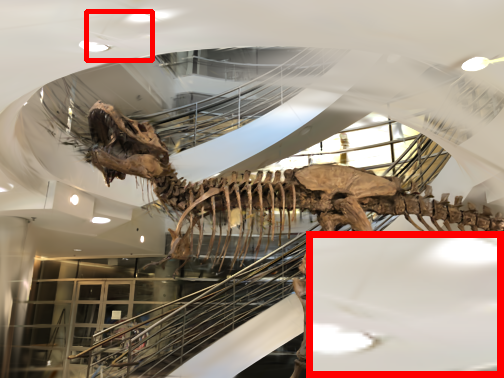}
    \end{minipage}
    \hfill
    \begin{minipage}{0.19\textwidth}
        \centering
        \text{DropGaussian} \\ \vspace{1mm}
        \includegraphics[width=\textwidth]{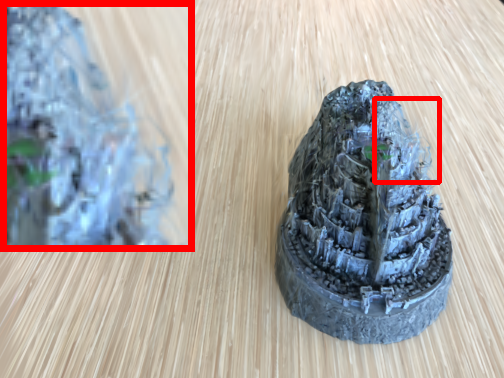} \\ \vspace{1mm}
        \includegraphics[width=\textwidth]{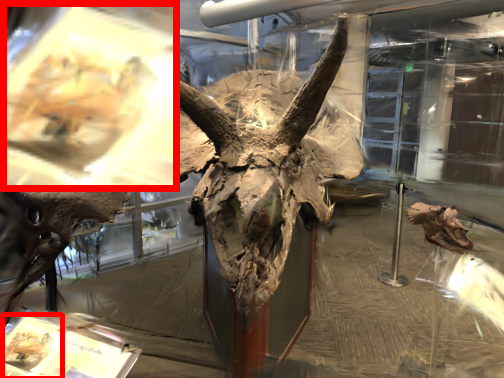} \\ \vspace{1mm}
        \includegraphics[width=\textwidth]{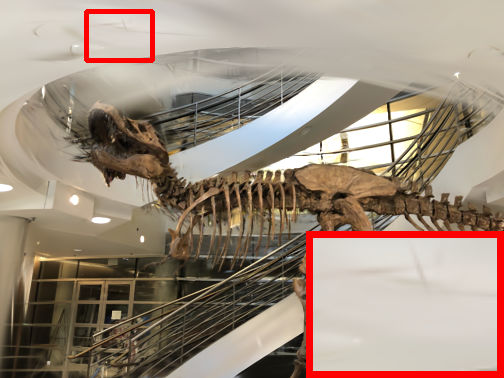}
    \end{minipage}
    \hfill
    \begin{minipage}{0.19\textwidth}
        \centering
        \textbf{DOC-GS (Ours)} \\ \vspace{1mm}
        \includegraphics[width=\textwidth]{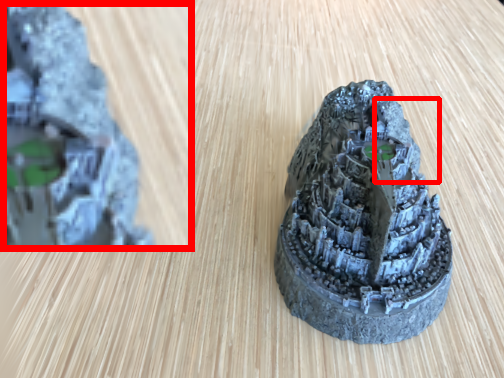} \\ \vspace{1mm}
        \includegraphics[width=\textwidth]{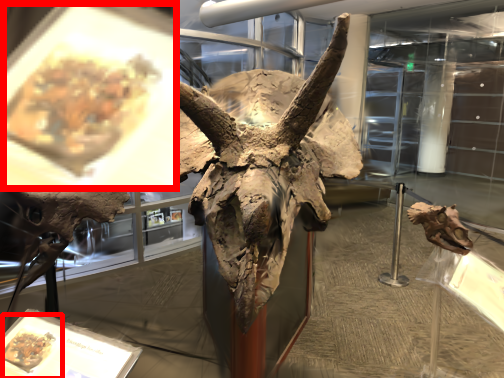} \\ \vspace{1mm}
        \includegraphics[width=\textwidth]{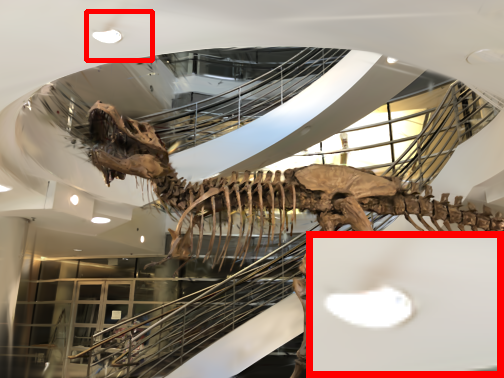}
    \end{minipage}
    \hfill
    \begin{minipage}{0.19\textwidth}
        \centering
        \text{Ground Truth} \\ \vspace{1mm}
        \includegraphics[width=\textwidth]{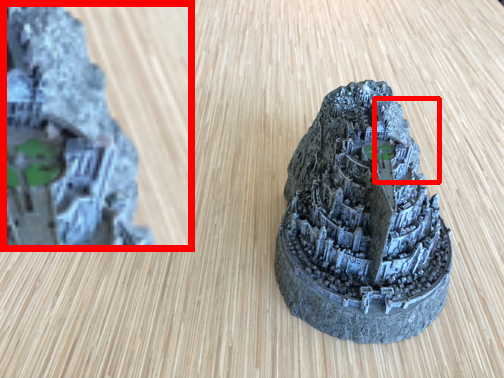} \\ \vspace{1mm}
        \includegraphics[width=\textwidth]{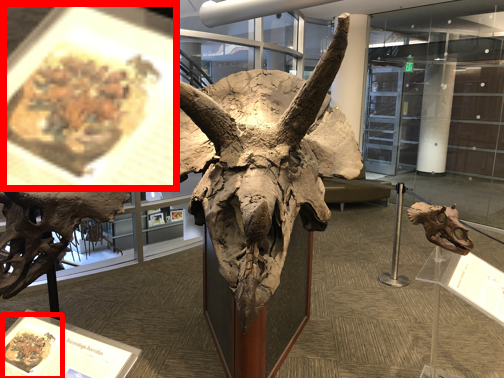} \\ \vspace{1mm}
        \includegraphics[width=\textwidth]{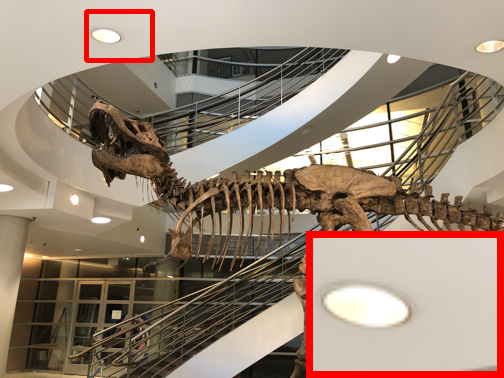}
    \end{minipage}
    \caption{\textbf{Qualitative comparison of our method and three existing methods on the LLFF~\cite{Mildenhall2019} dataset under 3-view settings.} Our method effectively suppresses floater artifacts and produces cleaner and more consistent geometric structures.}
    \label{fig_llff}
\end{figure*}


\begin{figure*}[!t]
    \centering
    \small 
    \begin{minipage}{0.19\textwidth}
        \centering
        \text{3DGS} \\ \vspace{1mm}
        \includegraphics[width=\textwidth]{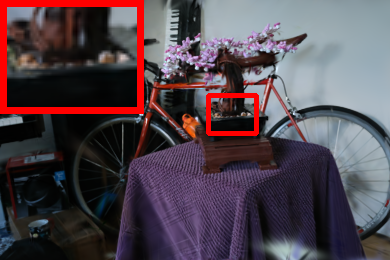} \\ \vspace{1mm}
        \includegraphics[width=\textwidth]{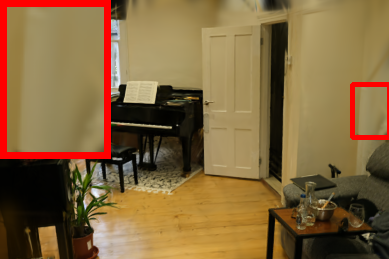} \\ \vspace{1mm}
        \includegraphics[width=\textwidth]{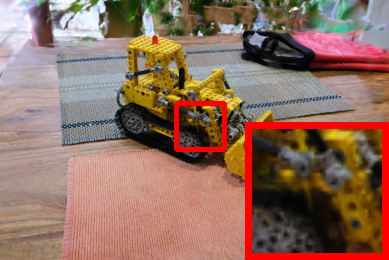}
    \end{minipage}
    \hfill
    \begin{minipage}{0.19\textwidth}
        \centering
        \text{DropoutGS} \\ \vspace{1mm}
        \includegraphics[width=\textwidth]{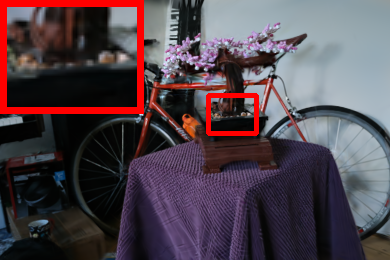} \\ \vspace{1mm}
        \includegraphics[width=\textwidth]{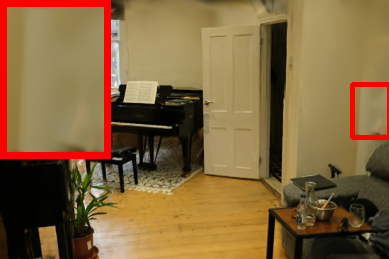} \\ \vspace{1mm}
        \includegraphics[width=\textwidth]{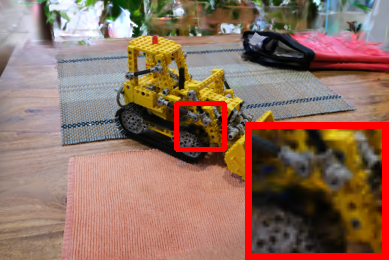}
    \end{minipage}
    \hfill
    \begin{minipage}{0.19\textwidth}
        \centering
        \text{DropGaussian} \\ \vspace{1mm}
        \includegraphics[width=\textwidth]{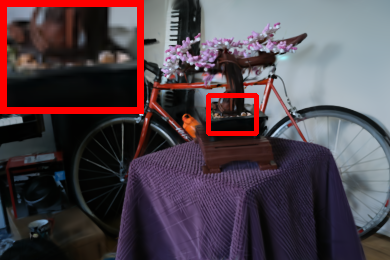} \\ \vspace{1mm}
        \includegraphics[width=\textwidth]{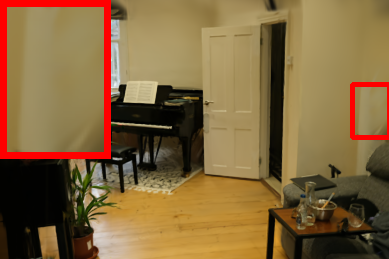} \\ \vspace{1mm}
        \includegraphics[width=\textwidth]{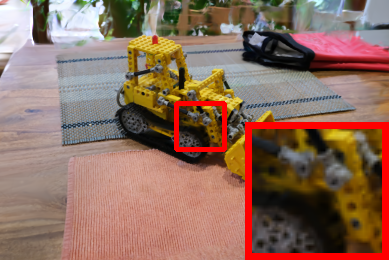}
    \end{minipage}
    \hfill
    \begin{minipage}{0.19\textwidth}
        \centering
        \textbf{DOC-GS (Ours)} \\ \vspace{1mm}
        \includegraphics[width=\textwidth]{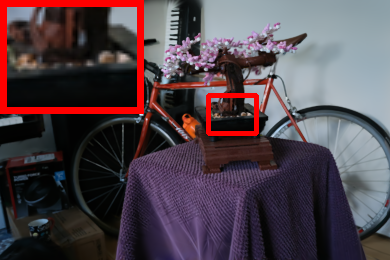} \\ \vspace{1mm}
        \includegraphics[width=\textwidth]{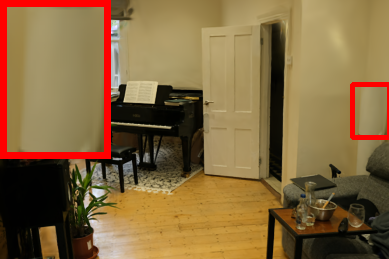} \\ \vspace{1mm}
        \includegraphics[width=\textwidth]{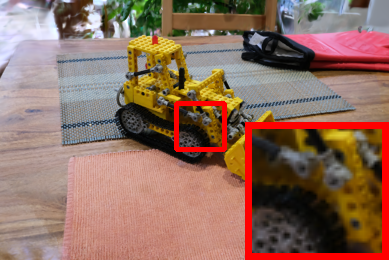}
    \end{minipage}
    \hfill
    \begin{minipage}{0.19\textwidth}
        \centering
        \text{Ground Truth} \\ \vspace{1mm}
        \includegraphics[width=\textwidth]{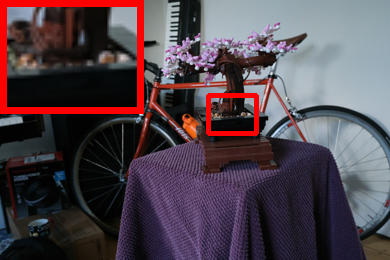} \\ \vspace{1mm}
        \includegraphics[width=\textwidth]{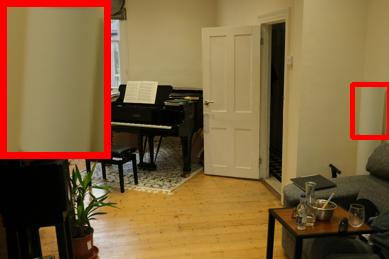} \\ \vspace{1mm}
        \includegraphics[width=\textwidth]{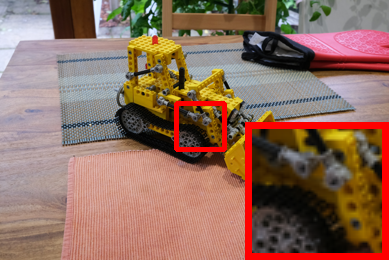}
    \end{minipage}
    \caption{\textbf{Qualitative comparison of our method and three existing methods on the MipNeRF360~\cite{Barron2022} dataset under 12-view settings.} Our method effectively suppresses floater artifacts and produces cleaner and more consistent geometric structures.
    }
    \label{fig:mipnerf360}
\end{figure*}

\subsection{Quantitative Comparison}

We quantitatively evaluate our proposed method against various baseline methods on the LLFF~\cite{Mildenhall2019}, MipNeRF-360~\cite{Barron2022} and Blender~\cite{NeRF} datasets under different sparse view conditions. Specifically, we use 3, 6, and 9 views for the LLFF dataset, 12 and 24 views for the MipNeRF-360 dataset and 8view for the Blender dataset. The corresponding results are presented in Tab.~\ref{tab:llff_results}, Tab.~\ref{tab:mipnerf360_results} and Tab.~\ref{tab:blender_results}, respectively. In particular, on the LLFF~\cite{Mildenhall2019} dataset,  our method achieved a PSNR of 21.38 dB under the highly restrictive 3-view setting, significantly outperforming 3DGS by 2.16 dB while also surpassing the recent work NexusGS~\cite{NexusGS} and DropGaussian~\cite{DropGaussian}. 
As the number of views increased, our method continue to demonstrate superior performance, the same conclusion holds true for the MipNeRF-360~\cite{Barron2022} and Blender~\cite{NeRF} dataset. This validates the effectiveness of our proposed dual-domain calibration framework. 
By jointly leveraging optimization-domain regularization and observation-domain anomaly feedback, our method effectively suppresses unreliable Gaussian primitives while preserving consistent scene structures. Furthermore, the results demonstrate that even under relatively dense-view settings, the proposed framework avoids over-pruning valid geometry, ensuring robust and high-fidelity reconstruction across varying degrees of data sparsity.

\subsection{Qualitative Comparison}

As illustrated in Fig.~\ref{fig_llff} and Fig.~\ref{fig:mipnerf360}, we present qualitative comparisons across diverse scenes, including 3DGS~\cite{3dgs}, DropGaussian~\cite{DropGaussian} , DropoutGS~\cite{DropoutGS}, ours and the ground truth (GT). Regions with noticeable differences are highlighted by red bounding boxes.
Compared to existing methods, our approach produces more coherent geometry and significantly suppresses haze-like floaters and structural distortions. In particular, stochastic dropout-based methods such as DropGaussian~\cite{DropGaussian} and DropoutGS~\cite{DropoutGS}  often introduce unstable structures or over-smooth details, while vanilla 3DGS suffers from noticeable floater artifacts under sparse views. In contrast, by jointly leveraging optimization-domain regularization and observation-domain anomaly feedback, our method effectively suppresses unreliable Gaussian primitives while preserving consistent scene structures.
As a result, our DOC-GS achieves sharper boundaries, cleaner surfaces, and more visually faithful reconstructions, demonstrating clear qualitative advantages over existing baselines.


\begin{table}[t]
\centering
\caption{Compatibility to existing sparse-view reconstruction methods with our design on the LLFF dataset (3-views). }
\label{tab:quantitative_results}
\setlength{\tabcolsep}{9pt}
\begin{tabular}{l|ccc}
\hline
\textbf{Methods} & \textbf{PSNR$\uparrow$} & \textbf{SSIM$\uparrow$} & \textbf{LPIPS$\downarrow$} \\
\hline
FSGS\cite{FSGS}& 20.43 & 0.682 &0.248 \\
FSGS + Ours          &20.69 & 0.705 & 0.213 \\
\hline
CoR-GS\cite{Cor-GS}        & 20.45 & 0.712 & 0.196 \\
CoR-GS + Ours        &20.72  & 0.719 &0.195  \\
\hline
DropGaussian\cite{DropGaussian}     & 20.76 &0.713 &0.200\\
DropGaussian + Ours    & 21.24 & 0.740 &  0.179\\
\hline
\end{tabular}
\label{tab:plug-and-play}
\end{table}

\begin{figure}[t]
\centering
\captionsetup[figure]{font=normal}
\captionsetup[subfigure]{labelformat=empty} 

\begin{subfigure}[t]{0.32\columnwidth}
    \centering
    \includegraphics[width=\textwidth]{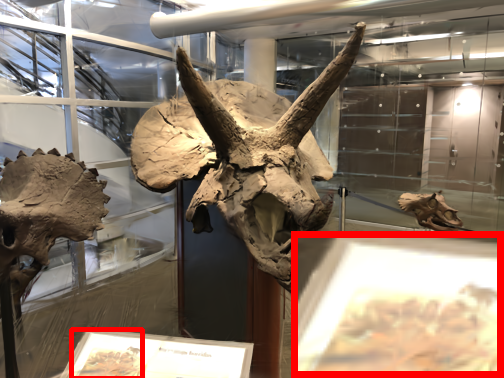}
    \caption{FSGS}
\end{subfigure}
\hfill
\begin{subfigure}[t]{0.32\columnwidth}
    \centering
    \includegraphics[width=\textwidth]{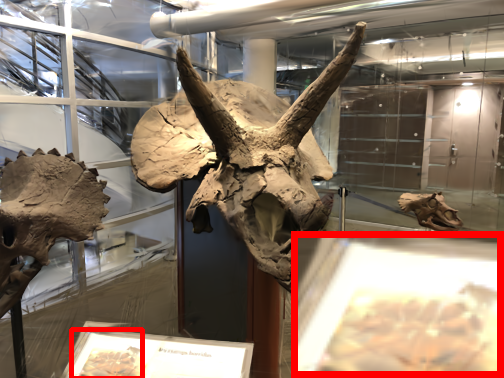}
    \caption{CoR-GS}
\end{subfigure}
\hfill
\begin{subfigure}[t]{0.32\columnwidth}
    \centering
    \includegraphics[width=\textwidth]{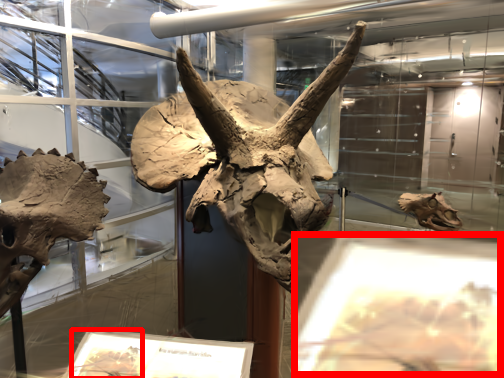}
    \caption{DropGaussian}
\end{subfigure}

\vspace{1mm}

\begin{subfigure}[t]{0.32\columnwidth}
    \centering
    \includegraphics[width=\textwidth]{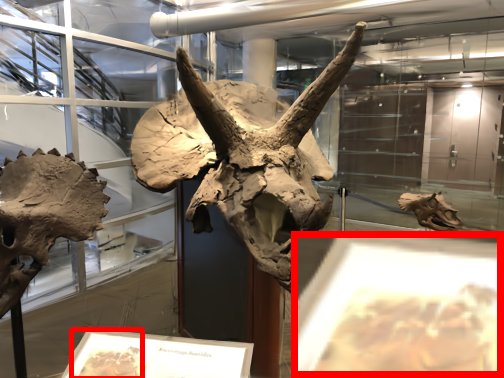}
    \caption{FSGS + Ours}
\end{subfigure}
\hfill
\begin{subfigure}[t]{0.32\columnwidth}
    \centering
    \includegraphics[width=\textwidth]{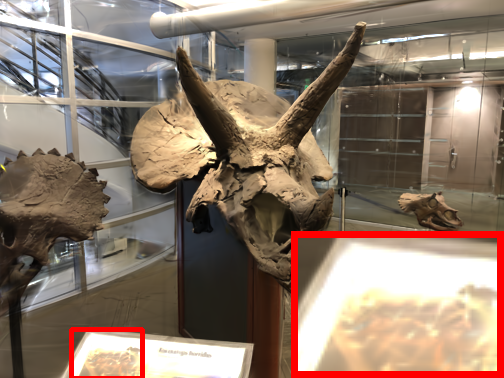}
    \caption{CoR-GS + Ours}
\end{subfigure}
\hfill
\begin{subfigure}[t]{0.32\columnwidth}
    \centering
    \includegraphics[width=\textwidth]{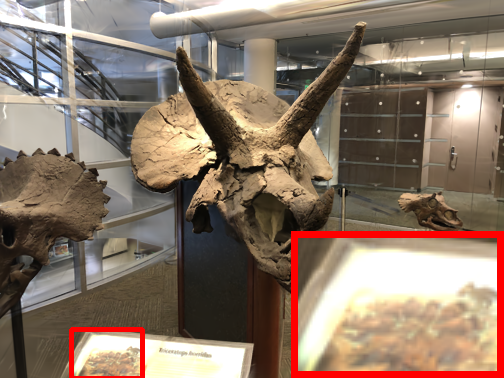}
    \caption{DropGaussian + Ours}
\end{subfigure}

\caption{Compatibility to 3DGS variants on the LLFF dataset (3-views). The  DOC-GS can be easily integrated into various 3DGS-based methods and improve their performance.}
\label{fig:plug-and-play}
\end{figure}

\subsection{Compatibility}

We present a lightweight and modular dual-domain calibration framework that can be seamlessly integrated into existing 3DGS variants. As shown in Tab.~\ref{tab:plug-and-play} and Fig.~\ref{fig:plug-and-play}, equipping representative methods such as FSGS~\cite{FSGS}, CoR-GS~\cite{Cor-GS}, and DropGaussian with our framework consistently leads to notable performance gains under sparse-view settings.
These results highlight the generality of our approach and demonstrate that the proposed dual-domain synergy effectively improves the reliability of Gaussian primitives, thereby enhancing the robustness and reconstruction fidelity of diverse 3DGS-based methods in highly under-constrained scenarios.

\subsection{Ablation Study}

To validate the contribution of each component, we conduct ablation experiments on the challenging LLFF 3-view setting. The quantitative results are summarized in Tab.~\ref{tab:ablation}. We adopt DropoutGaussian with random dropout as the baseline and a discrete depth-guided dropout strategy model as a strong reference.

\noindent\textbf{Effectiveness of optimization-domain regularization.}
We first evaluate the impact of replacing discrete depth-guided dropout with our Continuous Depth-Guided Dropout (CDGD). This modification improves PSNR from 20.74 dB (baseline) and 20.94 dB (DDGS) to 21.12 dB. The gain indicates that the proposed continuous formulation provides a smoother and more stable optimization process, effectively reducing abrupt masking effects and improving the consistency of Gaussian representations under sparse supervision.

\noindent\textbf{Effectiveness of observation-domain refinement.}
We then assess the proposed Dark Channel Prior-Guided Geometric Pruning (DCP-GP) independently. Applying DCP-GP to the baseline improves PSNR to 21.19 dB, demonstrating that observation-domain anomaly cues can effectively identify and suppress unreliable Gaussian primitives. This leads to the removal of haze-like floaters while preserving valid scene structures.

\noindent\textbf{Complementary dual-domain effect.}
Finally, combining CDGD and DCP-GP yields the best performance, achieving 21.38 dB PSNR. This confirms that the two components are complementary: the optimization-domain regularization improves the global stability of Gaussian learning, while observation-domain feedback further refines the representation by eliminating residual artifacts. Their synergy enables more reliable Gaussian primitives and results in superior reconstruction quality under sparse-view settings.

%


\begin{table}[!t]
    \centering
    \caption{Ablation study on the LLFF dataset (3 views). 
    }
    \label{tab:ablation}
    \begin{tabular}{c|c|c|ccc}
        \toprule
        DDGS & CDGD & DCP-GP & PSNR $\uparrow$ & SSIM $\uparrow$ & LPIPS $\downarrow$ \\
        \midrule
        $\times$ & $\times$ & $\times$ & 20.74 & 0.716 & 0.200 \\
        \checkmark & $\times$ & $\times$ & 20.94 & 0.722 & 0.196 \\
        $\times$ & \checkmark & $\times$ & \cellcolor{yellow!30}21.12 & \cellcolor{yellow!30}0.724 & \cellcolor{yellow!30}0.189 \\
        $\times$ & $\times$ & \checkmark & \cellcolor{orange!30}21.19 & \cellcolor{orange!30}0.733 & \cellcolor{orange!30}0.184 \\
        $\times$ & \checkmark & \checkmark & \cellcolor{red!30}21.38 & \cellcolor{red!30}0.748 & \cellcolor{red!30}0.176 \\
        \bottomrule
    \end{tabular}
\end{table}


\section{Conclusion}

In this paper, we revisit sparse-view 3D Gaussian splatting from a new perspective and formulate it as a dual-domain reliability inference problem, where the reliability of Gaussian primitives is inherently unobservable under limited supervision. 
Based on this formulation, we propose DOC-GS, a unified calibration framework that jointly leverages optimization-domain signals and observation-domain evidence to infer and refine Gaussian reliability. Specifically, we introduce a Continuous Depth-Guided Dropout (CDGD) to suppress weakly constrained primitives during optimization, and 
use the Dark Channel Prior (DCP) to detect structural inconsistencies in rendered images. These complementary cues are further integrated through a reliability-driven pruning mechanism, which progressively removes unreliable Gaussians.
Extensive experiments demonstrate that DOC-GS consistently improves reconstruction quality under all sparse-view settings compared with existing methods.






\bibliographystyle{ACM-Reference-Format}
\bibliography{sample-base}

\appendix

\end{document}